\documentclass[pdflatex,sn-mathphys-num]{sn-jnl}

\usepackage{graphicx}%
\usepackage{multirow}%
\usepackage{amsmath,amssymb,amsfonts}%
\usepackage{amsthm}%
\usepackage{mathrsfs}%
\usepackage[title]{appendix}%
\usepackage{xcolor}%
\usepackage{textcomp}%
\usepackage{manyfoot}%
\usepackage{booktabs}%
\usepackage{algorithm}%
\usepackage{algorithmicx}%
\usepackage{algpseudocode}%
\usepackage{listings}%

\usepackage{amsmath}
\usepackage{float}
\usepackage{booktabs}
\usepackage{rotating}
\usepackage{booktabs}
\usepackage{subcaption}

\theoremstyle{thmstyleone}%
\theoremstyle{thmstyletwo}%

\theoremstyle{thmstylethree}%

\begin{document}

\title[Article Title]{Revisiting the Ordering of Channel and Spatial Attention: A Comprehensive Study on Sequential and Parallel Designs}







\author[1]{\fnm{Zhongming} \sur{Liu}}\email{202226703019@jxnu.edu.cn}

\author[1]{\fnm{Bingbing}
\sur{Jiang}}\email{202226702041@jxnu.edu.cn}

\affil[1]{\orgdiv{School of Artificial Intelligence}, \orgname{Jiangxi Normal University}, \orgaddress{\city{Nanchang}, \postcode{330022}, \country{China}}}



\abstract{
Attention mechanisms have become a core component of deep learning models, with Channel Attention and Spatial Attention being the two most representative architectures. Current research on their fusion strategies primarily bifurcates into sequential and parallel paradigms, yet the selection process remains largely empirical, lacking systematic analysis and unified principles. We systematically compare channel-spatial attention combinations under a unified framework, building an evaluation suite of 18 topologies across four classes: sequential, parallel, multi-scale, and residual. Across two vision and nine medical datasets, we uncover a “data scale–method–performance” coupling law:
(1) in few-shot tasks ($N<1k$), the "Channel-Multi-scale Spatial" cascaded structure achieves optimal performance; 
(2) in medium-scale tasks ($1k \le N \le 50k$), parallel learnable fusion architectures demonstrate superior results;
(3) in large-scale tasks ($N>50k$), parallel structures with dynamic gating yield the best performance. Additionally, experiments indicate that the "Spatial→Channel" order is more stable and effective for fine-grained classification, while residual connections mitigate vanishing gradient problems across varying data scales. We thus propose scenario-based guidelines for building future attention modules. Code is open-sourced at \url{https://github.com/DWlzm}.
}




\maketitle

\section{Introduction}

Attention mechanisms have become a core component of deep learning models\cite{FENG2026132232,ELSAABRAHAM2025132379,ZHANG2025132338,WU2026105770}, particularly in Convolutional Neural Networks (CNNs), where they significantly enhance the performance of computer vision tasks by dynamically weighting key features and suppressing redundant information\cite{Transformer,LIU2025131369,ZHAO2025131101}. In the domain of visual attention, Channel Attention (CA) and Spatial Attention (SA) represent two core paradigms: Channel Attention focuses on "which feature channels to attend to," achieving feature recalibration by modeling inter-channel dependencies; Spatial Attention focuses on "which locations in the image to attend to," enhancing the model's spatial perception capability by strengthening discriminative regions.

With the deepening of research, single-dimensional attention mechanisms have gradually exposed their limitations: channel attention only focuses on the channel importance of global semantics but ignores the spatial structural information of feature maps; spatial attention can locate key regions but fails to distinguish the semantic value of different channels. For this reason, researchers have begun to explore fusion strategies for the two, giving rise to typical attention mechanisms such as CBAM\cite{CBAM} (channel-first then spatial) and BAM\cite{BAM} (parallel fusion of channel and spatial). However, existing fusion methods still have significant shortcomings: first, their design relies on empirical inspiration and lacks systematic theoretical support; second, they do not analyze the adaptive relationship between data scale, task type and fusion strategy, resulting in limited generalization ability of existing mechanisms. These gaps make the design of attention modules lack unified guidance, making it difficult to achieve optimal performance in diverse scenarios. To address the above problems, we sort out the design ideas of existing attention mechanisms\cite{VENTURINI2025105754,LU2026110288,LI2026108243,SI2025129866,FANG2025107474,ZHAO2025111520,CAI2025114285,WANG2025111907,11239364} and extract four core improvement directions for fusion strategies: sequential, parallel, multi-scale and residual.

This paper conducts a systematic re-examination of the combination modes of channel and spatial attention, with the core goal of answering three progressively in-depth key questions: (1) Does the combination mode of channel and spatial attention affect model performance? (2) Under different data scales and task types, what are the applicable conditions and advantage boundaries of sequential and parallel structures? (3) Is there an attention fusion paradigm that can adapt to diverse scenarios and provide universal design criteria for different tasks?

\begin{figure}[H]
    \centering
    \includegraphics[width=\linewidth]{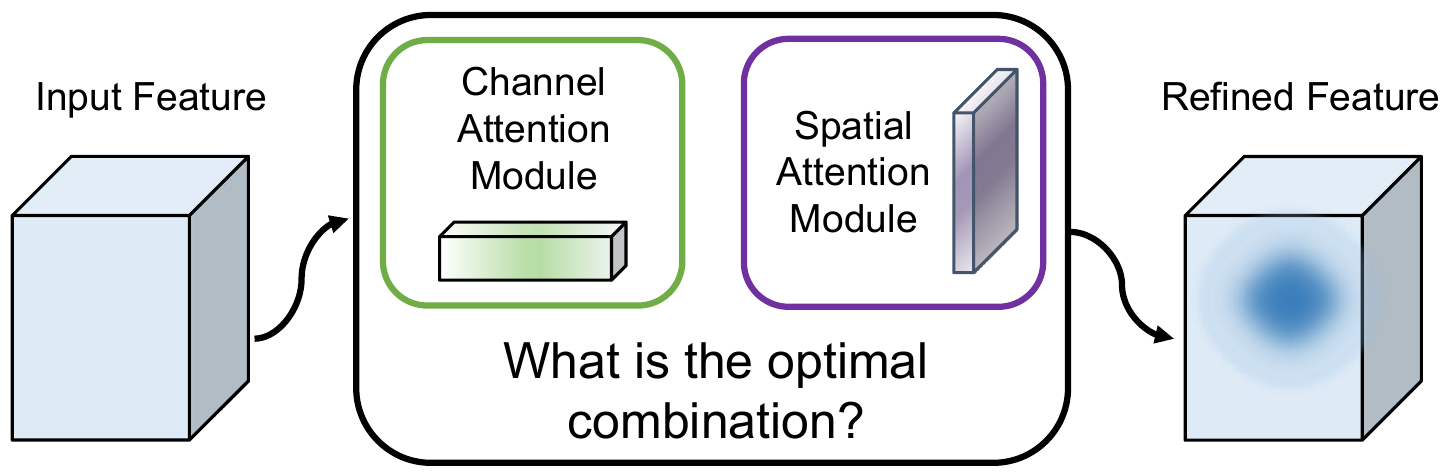}
    \caption{Searching for the optimal attention mechanism among various combinations of channel and spatial attention.}
    \label{fig:my_label}
\end{figure}

To answer the above questions, this paper conducts systematic experiments on multi-source datasets covering general vision and medical imaging, with the main contributions as follows:

\begin{enumerate}
    \item A systematic framework for combination strategies of channel-spatial attention is constructed, covering 18 topological structures across four categories (sequential, parallel, multi-scale, and residual connection), which fills the gap of the lack of comprehensive comparison in existing research;
    \item The coupling mechanism of "data scale - combination mode - model performance" is revealed. The impacts of different attention orders on feature distribution and gradient flow are verified through visualization and statistical analysis, and the advantage principle of spatial-first order in fine-grained classification is clarified;
    \item Scene-adaptive attention design criteria are proposed: the "channel-multi-scale spatial" cascaded structure is preferred for few-shot scenarios, the parallel learnable fusion strategy is adopted for medium-shot scenarios, and the dynamic gated parallel structure is introduced for large-shot scenarios, providing direct references for the design of subsequent attention modules.
    \item We systematically proposed for the first time more than ten novel attention fusion mechanisms such as GC$\&$SA$^2$ and TGPFA. Through innovative designs including learnable weights, dynamic gating and multi-scale cascading, we enriched the design space of attention topological structures, providing plug-and-play module options for different data scales and task types.
\end{enumerate}

The subsequent structure of this paper is arranged as follows: Section 2 reviews relevant research on channel attention, spatial attention, and joint attention; Section 3 elaborates on the design principles of various attention combination strategies; Section 4 verifies the effectiveness of the design through experiments on multiple datasets; Section 5 discusses the core laws discovered from the experiments; Section 6 summarizes the full paper and points out future research directions.

\section{Related Work}

\subsection{Channel Attention Mechanisms}
The core idea of Channel Attention (CA) is to model the interdependencies between channels and achieve the recalibration of feature responses through adaptive weighting of features from different channels. The pioneering work SENet~\cite{SENet} first proposed the "squeeze–excitation" mechanism, which extracts channel-level global information via global average pooling and learns the importance weights of each channel, thereby significantly enhancing the representational capacity of convolutional neural networks. Building on this, subsequent studies have improved channel attention from perspectives such as efficiency, structure, and context modeling. For instance, ECA-Net~\cite{ECANet} replaces fully connected layers with one-dimensional convolutions to avoid information loss caused by dimensionality reduction operations; SKNet~\cite{SKNet} proposes a multi-scale mechanism with selectable convolution kernels to realize adaptive fusion of features from different receptive fields; GCNet~\cite{GCNet} integrates global context modeling into channel attention, balancing global information aggregation and computational efficiency; and Coordinate Attention~\cite{CoordinateAttention} explicitly introduces spatial coordinate information into channel modeling, enabling attention to be sensitive to positional information while maintaining global dependencies.

Although channel attention has achieved remarkable results in enhancing the feature expression of networks, it mainly focuses on the global semantic importance of the channel dimension and often ignores the structural information of feature maps in the spatial dimension. Therefore, researchers have begun to study spatial attention mechanisms.

\subsection{Spatial Attention Mechanisms}
Spatial Attention (SA) aims to model the saliency differences of features in the spatial dimension, enabling the model to focus on more discriminative regions in images. Unlike channel attention, spatial attention emphasizes "where to focus" by assigning weights to spatial positions. Methods such as PSA (Parallel Spatial Attention)~\cite{PSA} and Spatial Group-wise Enhance (SGE)~\cite{li2019sge} have further improved the modeling approach of spatial attention: PSA enhances the flexibility of spatial dependency modeling through parallel multi-scale feature aggregation, while SGE adopts an intra-group normalization strategy to capture salient spatial structural information in a lightweight manner. In addition, Coordinate Attention~\cite{CoordinateAttention} is also regarded as an extended form that introduces coordinate encoding in the spatial dimension, allowing the network to balance long-range dependencies and positional information.

However, relying solely on spatial attention also has certain limitations. It usually focuses on salient spatial regions in feature maps but fails to distinguish the semantic importance carried by different channels, thus making it difficult to fully utilize complementary information between channels. Furthermore, the modeling of spatial attention often relies on local convolution or aggregation operations, with limited ability to capture global dependencies. To further enhance the representational capacity of models, researchers generally choose to combine spatial attention with channel attention to achieve synergistic enhancement in both semantic and spatial dimensions. Based on this, different studies have proposed various fusion methods such as sequential and parallel, which is exactly the core issue to be explored in depth in this paper: the impact of combination strategies of channel and spatial attention on feature modeling and the resulting performance differences.

\subsection{Joint Channel–Spatial Attention Mechanisms}

\subsubsection{Sequential Designs}

CBAM\cite{CBAM} first proposed the combination of channel attention and spatial attention, adopting a sequential structure of "channel attention first, then spatial attention". By sequentially weighting feature maps with channel attention and spatial attention, this method enables the model to adaptively adjust feature representations at two levels: "which features to focus on" and "which positions to focus on", thereby enhancing the network's ability to represent key targets\cite{LV2026109537,JI2026113141,MA2025,PENG2026108807}. Subsequent studies have optimized this sequential structure: Chen et al.\cite{CHEN2025111958} introduced median pooling into the channel attention of CBAM to improve feature expression capability; Peng et al.\cite{GraphConvolutionAttention} proposed Graph Convolution Attention, which embeds a graph convolution module into the channel attention of CBAM to enhance the ability to model complex channel dependencies; Rahman et al.\cite{Rahman_2024_CVPR} introduced a multi-scale convolutional attention mechanism after the CBAM framework to achieve adaptive modeling of multi-scale features; Wazir et al.\cite{Wazir} improved the spatial attention on the basis of CBAM, fusing standardized pooling and min pooling to enhance the information richness of spatial feature maps.
The sequential connection of spatial attention and channel attention can effectively decouple the modeling tasks of "what" and "where", but their order has always relied on experience and lacks scientific basis.

\subsubsection{Parallel Designs}

Parallel attention mechanisms capture multi-dimensional contextual information more fully by modeling channel and spatial features simultaneously. The Bottleneck Attention Module (BAM) proposed by Park et al.\cite{BAM} introduces parallel channel attention and spatial attention branches into the residual bottleneck structure to adaptively recalibrate feature maps, thereby enhancing feature expression capability. The Dual Attention Network (DANet) proposed by Fu et al.\cite{DANet} designs a Position Attention Module (PAM) and a Channel Attention Module (CAM), which model long-range dependencies in the spatial and channel dimensions respectively to adaptively fuse local features and global semantics. On this basis, subsequent studies have enriched the parallel structure: Ma et al.\cite{GCE} proposed the GCE module, which effectively reduces computational complexity and strengthens local feature expression by grouping feature maps and combining with the CBAM attention mechanism; Pramanik et al.\cite{PCBAM} proposed the PCBAM module, which fuses CBAM with Position Attention (PAM) to improve the ability to model local spatial relationships; the CSAM module by Lei et al.\cite{CS-Net} extracts more refined global information through parallel channel self-attention and spatial self-attention mechanisms, and performs prominently in medical image segmentation tasks.

\section{Method}
\label{sec:method}
This section first introduces the core structures, mathematical expressions, and functional positioning of three types of basic attention components (Channel Attention, Spatial Attention, and Gate Attention); on this basis, it systematically elaborates four fusion paradigms of Channel Attention and Spatial Attention, including sequential mode, parallel mode, multi-scale information mode, and residual connection mode, presenting the design logic and implementation details of each combination strategy.

\begin{figure*}
    \centering
    \includegraphics[width=\linewidth]{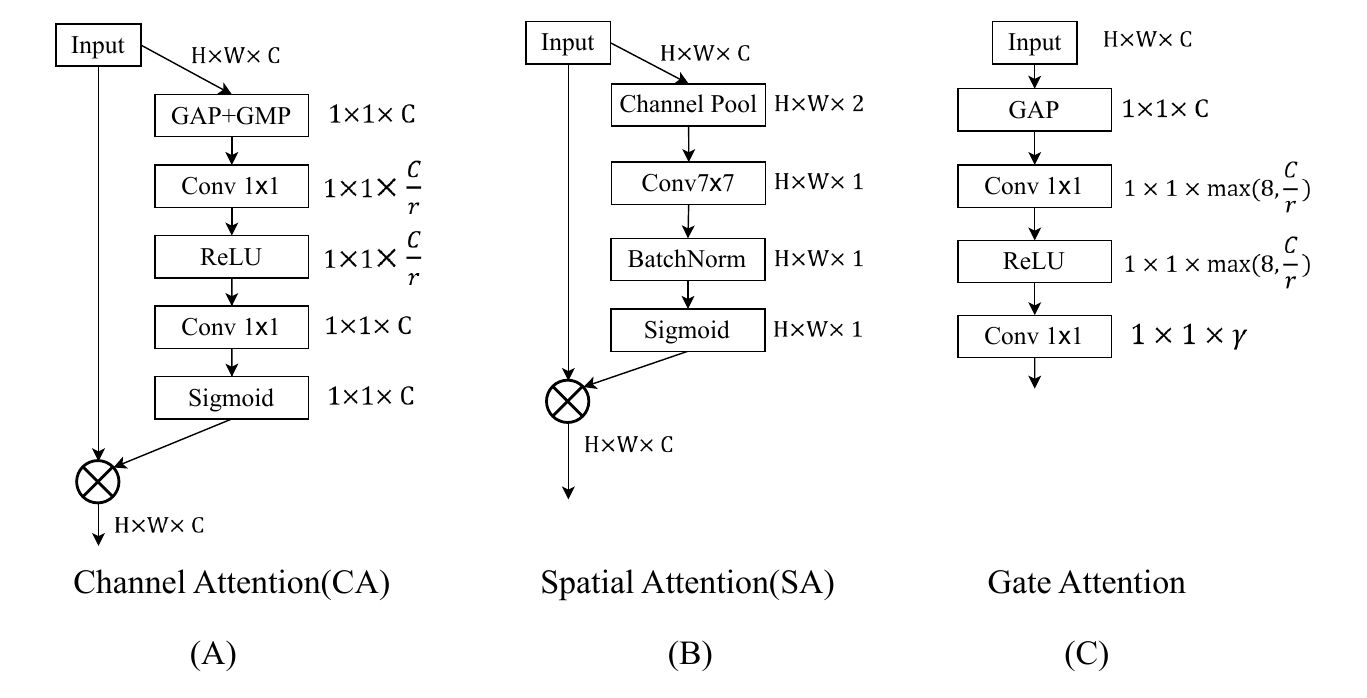}
    \caption{Structure Diagram of Basic Components}
    \label{fig:BaseModule}
\end{figure*}

\subsection{Preliminaries}

As shown in Figure~\ref{fig:BaseModule}(A),
\textbf{Channel Attention (CA)} first performs global average pooling and global max pooling on the input feature $\mathbf{X}_{CA}^{in} \in \mathbf{R}^{H \times W \times C}$ along the spatial dimension to extract two types of global statistical descriptors $\{ \mathbf{V}_{CA}^{1} , \mathbf{V}_{CA}^{2} \} \in \mathbf{R}^{1 \times 1 \times C}$:
\begin{equation}
    \mathbf{V}_{CA}^{1} = \text{GAP}(\textbf{X}_{CA}^{in}) ,  \mathbf{V}_{CA}^{2} = \text{GMP}(\textbf{X}_{CA}^{in}).
\end{equation}
Subsequently, a two-layer perceptron with shared weights is used to process the two pooling results respectively, and after summing them, a sigmoid activation function is applied to generate the channel attention weights $\mathbf{W}_{CA} \in \mathbf{R}^{1 \times 1 \times C}$:
\begin{equation}
    \text{MLP}(r,x) = \text{Conv}_{1\times 1, \frac{C}{r} \rightarrow C } \big( \text{ReLU} ( \text{Conv}_{1\times 1, C \rightarrow \frac{C}{r} } ( x ) ) \big),
\end{equation}
\begin{equation}
\mathbf{W}_{CA} = \text{Sigmoid} \!\left( \mathrm{MLP}_{shared} \!\left( \mathbf{V}_{CA}^{1} \right) + \mathrm{MLP}_{shared} \!\left( \mathbf{V}_{CA}^{2} \right) \right) .
\end{equation}
Finally, the weights are multiplied with the original features channel-wise to enhance important channels and suppress redundant channels:
\begin{equation}
   \mathbf{X}_{CA}^{out} = \mathbf{W}_{CA} \odot 
\mathbf{X}_{CA}^{in}.
\end{equation}
Through this design, the model can focus on learning the interdependencies between channels, thereby enhancing its ability to express category semantics.
For the convenience of subsequent paper writing, we formulate Channel Attention as follows:
\begin{equation}
    \mathbf{X}_{CA}^{out} = \text{CA}_{ r = 8 }(\mathbf{X}_{CA}^{in}) ,
\end{equation}
where $r$ denotes the compression ratio of the MLP.

As shown in Figure~\ref{fig:BaseModule}(B),
\textbf{Spatial Attention (SA)} first performs average pooling and max pooling on the input feature $\textbf{X}_{SA}^{in} \in \mathbf{R}^{H \times W \times C}$ along the channel dimension respectively, yielding two single-channel statistical maps $\{ \mathbf{V}_{SA}^{1} , \mathbf{V}_{SA}^{2} \} \in \mathbf{R}^{H \times W \times 1}$:
\begin{equation}
    \mathbf{V}_{SA}^{1} = \text{GAP}(\textbf{X}_{SA}^{in}) ,  \mathbf{V}_{SA}^{2} = \text{GMP}(\textbf{X}_{SA}^{in}).
\end{equation}
Subsequently, the two statistical maps are concatenated along the channel dimension to form a two-channel feature map; then, a convolution operation is used to learn the importance of spatial positions, and a sigmoid activation function is applied to generate the spatial attention weights $\mathbf{W}_{SA} \in \mathbf{R}^{H \times W \times 1}$:
\begin{equation}
    \mathbf{W}_{SA} = \text{Sigmoid}\!\left( \mathrm{Conv}_{7\times7}\!\left( \text{Concat}  [\mathbf{V}_{SA}^{1},\; \mathbf{V}_{SA}^{2}]  \right) \right).
\end{equation}
Finally, the weights are multiplied with the original features position-wise to enhance important spatial positions and suppress background regions:
\begin{equation}
    \mathbf{X}_{SA}^{out} = \mathbf{W}_{SA} 
    \odot 
\mathbf{X}_{in}^{in}.
\end{equation}
Through this design, the model can focus on learning the saliency of spatial positions, thereby enhancing its ability to express textures, edges, and target localization. For the convenience of subsequent paper writing, Spatial Attention is formulated as follows:
\begin{equation}
    \mathbf{X}_{SA}^{out} = \text{SA}_{c \times c} ( \mathbf{X}_{SA}^{in} ) ,
\end{equation}
where $c$ denotes the kernel size of the convolution for extracting local spatial features.

As shown in Figure~\ref{fig:BaseModule}(C),
\textbf{Gate Attention (GA)} first performs global average pooling on the input feature $\mathbf{X}_{GA}^{in} \in \mathbf{R}^{H \times W \times C}$ along the spatial dimension, compressing the $H\times W \times C$ tensor into a $1\times 1 \times C$ channel descriptor vector $\mathbf{V}_{GA}$:
 \begin{equation}
     \mathbf{V}_{GA} = \text{GAP} ( \mathbf{X}_{GA}^{in} ).
 \end{equation}
 Subsequently, two dimensionality reduction operations are performed using a multi-layer perceptron composed of two $1\times 1$ convolutions and ReLU to extract compact interdependencies between channels:
 \begin{equation}
     {l}_{GA} = \text{Conv}_{1 \times 1 , \frac{C}{r} \rightarrow 1 } ( \text{ReLU}( \text{Conv}_{1 \times 1 , C \rightarrow \frac{C}{r} } ( \mathbf{V}_{GA} ) ) ).
 \end{equation}
 Finally, a $1\times 1\times \gamma$ gating scalar is output through the Sigmoid activation function to perform global scaling on the original features as a whole:
 \begin{equation}
     \mathbf{X}_{GA}^{out} = \text{Sigmoid} ( {l}_{GA} ) \odot \mathbf{X}_{GA}^{in}.
 \end{equation}
 Through this design, the model acquires global context awareness at an extremely low computational cost, achieving adaptive intensity adjustment of feature maps instead of fine-grained channel-wise or position-wise weighting, thus balancing efficiency and expressiveness. GA can serve as a gate to modulate other attention mechanisms. For the convenience of subsequent paper writing, we formulate Gate Attention as follows:
 \begin{equation}
      \mathbf{X}_{GA}^{out} = GA(  \mathbf{X}_{GA}^{in}  ).
 \end{equation}

\subsection{Sequential Mode}
\begin{figure}[H]
    \centering
    \includegraphics[width=0.7\linewidth]{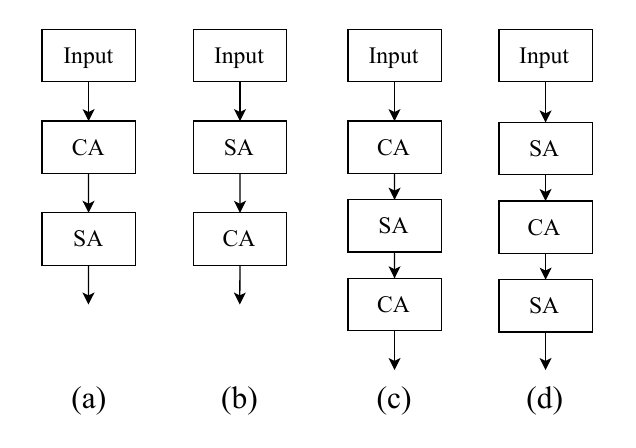}
    \caption{Structure Diagram of Sequential Mode}
    \label{fig1}
\end{figure}

\textbf{Channel-Spatial Attention (CSA)} CSA, also known as CBAM  (Convolutional Block Attention Module), adopts a serial structure (Figure~\ref{fig1}(a)): it first uses channel attention to select "which channels are worth focusing on", then employs spatial attention to determine "where to focus", with these two steps acting on the input feature map $\mathbf{X}^{\mathrm{in}}_{(a)}$ in sequence to achieve layer-by-layer refinement of "important channels—important positions". It is formulated as follows:
\begin{equation}
    \mathbf{X}_{(a)}^{out}= \text{SA}  \text{(CA}(\textbf{X}_{(a)}^{in})).
\end{equation}

\textbf{Spatial–Channel Attention (SCA)} SCA also adopts a serial structure (Figure~\ref{fig1}(b)): it first uses spatial attention to demarcate "where to focus", then employs channel attention to decide "which channels are worth focusing on". These two steps act on the input feature map $\mathbf{X}^{\mathrm{in}}_{(b)}$ in sequence, ultimately achieving progressive recalibration of "important positions—important channels". It is formulated as follows:
\begin{equation}
    \mathbf{X}_{(b)}^{out} = \text{CA}  \text{(SA}(\textbf{X}_{(b)}^{in})).
\end{equation}

\textbf{Channel-Spatial-Channel Attention (CSCA)} It adopts a three-stage serial structure (Figure~\ref{fig1}(c)), applying channel attention, spatial attention, and channel attention in sequence. This method first processes the input features with channel attention to generate the first layer of channel weights and complete channel enhancement; then feeds the channel-enhanced features into the spatial attention module to learn the importance of spatial positions and complete spatial enhancement; finally applies channel attention again to the spatial-enhanced features to generate the second layer of channel weights, further strengthening important channels and suppressing redundant channels. It is formulated as follows:
\begin{equation}
    \mathbf{X}_{(c)}^{out} = \text{CA(SA}  \text{(CA}(\textbf{X}_{(c)}^{in}))).
\end{equation}

\textbf{Spatial-Channel-Spatial Attention (SCSA)} SCSA adopts a three-stage serial structure (Figure~\ref{fig1}(d)), applying spatial attention, channel attention, and spatial attention in sequence. This method first processes the input features with spatial attention to generate the first layer of spatial weights and complete spatial enhancement; then feeds the spatial-enhanced features into the channel attention module to learn the interdependencies between channels and complete channel enhancement; finally applies spatial attention again to the channel-enhanced features to generate the second layer of spatial weights and further refine spatial saliency. It is formulated as follows:
\begin{equation}
    \mathbf{X}_{(d)}^{out} = \text{SA(CA}  \text{(SA}(\textbf{X}_{(d)}^{in}))).
\end{equation}

\subsection{Parallel Mode}

\begin{figure}[H]
    \centering
    \includegraphics[width=\linewidth]{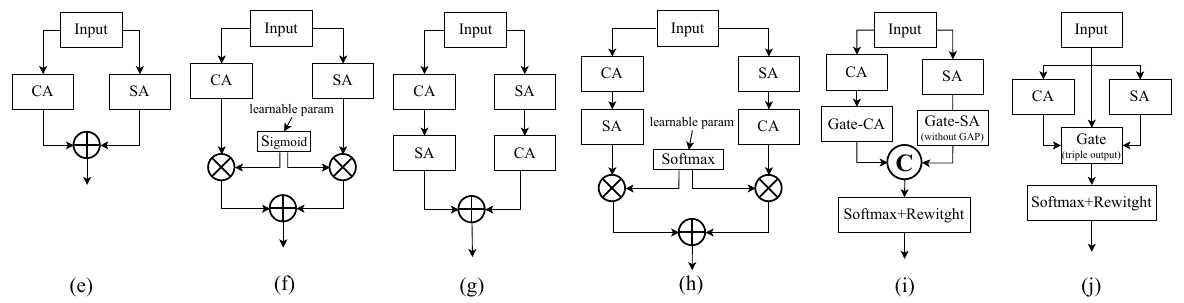}
    \caption{Structure Diagram of Parallel Mode}
    \label{fig2}
\end{figure}

\textbf{Channel $\&$ Spatial Additive Attention (C$\&$SA$^2$)}
It adopts a parallel structure, as shown in Figure~\ref{fig2}(e). With this parallel design, the model can simultaneously process information from both the channel and spatial dimensions, avoiding the limitation in serial structures where spatial attention relies on the output of channel attention. It is formulated as follows:
\begin{equation}
    \mathbf{X}_{(e)}^{out} = \text{SA}(\textbf{X}_{(e)}^{in} ) + \text{CA}(\textbf{X}_{(e)}^{in} ).
\end{equation}

\textbf{Channel $\&$ Spatial Adaptive-Fusion Attention (C$\&$SAFA)} It adopts a dual-branch parallel structure, as shown in Figure \ref{fig2}(f). This method first computes channel attention-enhanced features and spatial attention-enhanced features in parallel, then introduces learnable scalar parameters as gating units, normalizes them into dynamic weights through the Sigmoid activation function, and finally performs weighted fusion of the channel-enhanced features and spatial-enhanced features. It is formulated as follows:
\begin{equation}
 \mathbf{W}_{(f)} = \text{Sigmoid}\text{(Learnable Parameter) },
\end{equation}
\begin{equation}
    \mathbf{X}_{(f)}^{out} = \mathbf{W}_{(f)} \odot  \text{CA}(\textbf{X}_{(f)}^{in}) +( 1 - \mathbf{W}_{(f)} ) \odot  \text{SA}(\textbf{X}_{(f)}^{in}) .
\end{equation}

\textbf{Bidirectional Channel-Spatial Attention (Bi-CSA)}
As shown in Figure~\ref{fig2}(g), this method first constructs two parallel branches: the first branch applies channel attention and spatial attention in sequence, while the second branch applies spatial attention and channel attention in sequence; then the outputs of the two branches are directly added element-wise to obtain the final enhanced features. It is formulated as follows:
\begin{equation}
    \mathbf{X}_{(g)}^{out} = \text{SA}  \text{(CA}(\textbf{X}_{(g)}^{in})) + \text{CA}  \text{(SA}(\textbf{X}_{(g)}^{in})).
\end{equation}

\textbf{Bidirectional Channel-Spatial Adaptive-Fusion Attention (Bi-CSAFA)}
It adopts a dual-branch parallel structure, as shown in Figure \ref{fig2}(h). This method also constructs two parallel branches: the first branch sequentially applies channel attention and spatial attention, while the second branch sequentially applies spatial attention and channel attention; then, learnable fusion weights are calculated for each branch and normalized using the softmax function to ensure the sum of the two weights is 1; finally, the outputs of the two branches are weighted and fused according to the normalized weights. Through this dual-branch adaptive design, the model can automatically select a more favorable processing order based on the characteristics of input features, thereby achieving more robust performance under different data patterns. It is formulated as follows:
\begin{equation}
    \text{Softmax}(\mathbf{z}_i) 
  = \frac{e^{z_i}}{\sum_{j=1}^{K} e^{z_j}},
  \quad i = 1,2,\dots,K
\end{equation}
\begin{equation}
    \{ \mathbf{W}^{1}_{(h)} , \mathbf{W}^{2}_{(h  )} \} = \text{Softmax}\text{(Learnable Parameter) },
\end{equation}
\begin{equation}
    \mathbf{X}_{(h)}^{out} = \mathbf{W}^{1}_{(h)} \odot \text{SA}  \text{(CA}(\textbf{X}_{(h)}^{in})) +\mathbf{W}^{2}_{(h)} \odot  \text{CA} \text{(SA}(\textbf{X}_{(h)}^{in})) .
\end{equation}

\textbf{Gated Channel $\&$ Spatial Additive Attention (GC$\&$SA$^2$)}
It applies channel attention and spatial attention simultaneously and performs adaptive fusion via a dynamic gating mechanism. As shown in Figure \ref{fig2}(i), this method first computes channel attention and spatial attention in parallel on the same input feature to obtain channel-enhanced features and spatial-enhanced features, respectively; then constructs two lightweight gating heads: the channel gating head generates scalar gating values in the channel dimension, and the spatial gating head generates scalar gating values in the spatial dimension; next, concatenates the two gating values and normalizes them using the softmax function to obtain adaptive fusion weights for the two branches; finally, performs weighted fusion of the channel-enhanced features and spatial-enhanced features according to the normalized weights. Through this dynamic gating design, the model can adaptively adjust the relative importance of the channel and spatial dimensions based on input features, achieving better feature representation capability with almost no additional computational overhead. It is formulated as follows:
\begin{equation}
    \mathbf{X}^{CA}_{(i)} = \text{CA}(\textbf{X}_{(i)}^{in}),    \mathbf{X}^{SA}_{(i)} = \text{SA}(\textbf{X}_{(i)}^{in}),
\end{equation}
\begin{equation}
\begin{split}
    \{\mathbf{W}^{1}_{(i)},\mathbf{W}^{2}_{(i)} \} =  
    \text{Softmax}\bigl( \text{Concat}\bigl[ &\text{Gate-CA}(\mathbf{X}^{CA}_{(i)}), \\
    &\text{Gate-SA}(\mathbf{X}^{CA}_{(i)}) \bigr] \bigr),
\end{split}
\end{equation}
\begin{equation}
    \mathbf{X}_{(i)}^{out} =  \mathbf{W}^{1}_{(i)} \odot \mathbf{X}^{CA}_{(i)} + \mathbf{W}^{2}_{(i)} \odot \mathbf{X}^{SA}_{(i)}.
\end{equation}
Among them, Gate-CA refers to GA, and Gate-SA is the result of removing global average pooling from GA.

\textbf{Triple-Gate Parallel Fusion Attention (TGPFA)}
It adopts a triple parallel structure, as shown in Figure \ref{fig2}(j). This method first computes the original features, channel attention-enhanced features, and spatial attention-enhanced features respectively from the same input feature; then constructs a gating network to learn adaptive fusion weights for the three branches; next normalizes the three weights using the softmax function to ensure the sum of the three weights is 1; finally performs weighted fusion of the original features, channel-enhanced features, and spatial-enhanced features according to the normalized weights. Through this triple parallel adaptive design, the model can simultaneously retain original information, leverage the advantages of channel enhancement, and utilize the advantages of spatial enhancement, thereby achieving more stable and robust feature representation in complex scenarios. It is formulated as follows:
\begin{equation}
    \mathbf{X}_{(j)}^{CA} =  \text{CA}(\textbf{X}_{(j)}^{in}),
    \mathbf{X}_{(j)}^{SA} =  \text{SA}(\textbf{X}_{(j)}^{in}),
\end{equation}
\begin{equation}
\begin{split}
    \bigl\{ \mathbf{W}_{(j)}^{1} , \mathbf{W}_{(j)}^{2} , \mathbf{W}_{(j)}^{3} \bigr\} 
    &= \text{Softmax}\bigl( \text{Gate}\bigl( \text{CA}(\mathbf{X}_{(j)}^{in}), \\
    &\quad \text{SA}(\mathbf{X}_{(j)}^{in}), \mathbf{X}_{(j)}^{in} \bigr) \bigr),
\end{split}
\end{equation}
\begin{equation}
     \mathbf{X}_{(j)}^{out} = 
     \mathbf{W}_{(j)}^{1} \odot  \textbf{X}_{(j)}^{in} + 
     \mathbf{W}_{(j)}^{2} \odot  \mathbf{X}_{(j)}^{CA} + 
     \mathbf{W}_{(j)}^{3} \odot  \mathbf{X}_{(j)}^{SA}.
\end{equation}

\subsection{Residual Connection Mode}
\begin{figure}[H]
    \centering
    \includegraphics[width=0.8\linewidth]{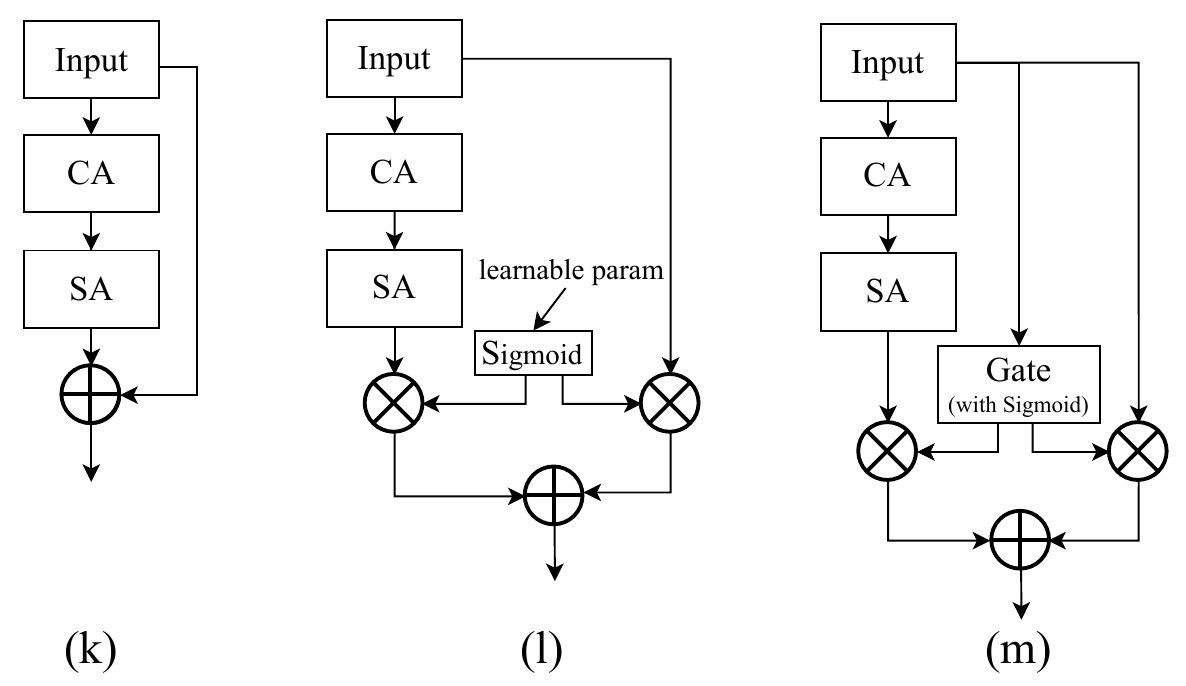}
    \caption{Structure Diagram of Residual Connection Pattern}
    \label{fig4}
\end{figure}

\textbf{Residual Channel-Spatial Attention (RCSA)}
An identity shortcut is added after CSA, and the attention output is summed with the original input (Figure \ref{fig4}(k)). This addition transforms the "multiplicative gating" into "additive correction": when the attention weights tend to 0, both gradients and features can be transmitted losslessly along the identity path, thereby preserving low-frequency background information, preventing excessive suppression, and alleviating early gradient vanishing. It is formulated as follows:
\begin{equation}
    \mathbf{X}_{(k)}^{out} = \textbf{X}_{(k)}^{in} + \text{SA}(\text{CA}(\textbf{X}_{(k)}^{in})).
\end{equation}

\textbf{Adaptive Residual Channel-Spatial Attention } 
\textbf{(ARCSA)}
It adopts an adaptive residual connection structure and introduces learnable fusion weights for the CSA-enhanced features and the original input. As shown in Figure \ref{fig4}(l), this method sequentially applies channel attention and spatial attention to the input features in the manner of CSA to generate attention-enhanced features; then introduces a learnable residual weight parameter to perform weighted fusion of the original input and the attention-enhanced features according to adaptive weights, where the weight of the original input is 1 minus the residual weight, and the weight of the attention-enhanced features is the residual weight itself. Through this adaptive residual design, the model can automatically adjust the relative contributions of original information and attention-enhanced information according to the training process and data characteristics. It is formulated as follows:
\begin{equation}
    W_{(l)} = \text{Sigmoid}\text{(Learnable Parameter) },
\end{equation}
\begin{equation}
    \mathbf{X}_{(l)}^{out} = \textbf{X}_{(l)}^{in} \odot (1-W_{(l)}) + \text{SA}(\text{CA}(\textbf{X}_{(l)}^{in})) \odot W_{(l)} .
\end{equation}

\textbf{Gated Residual Channel–Spatial Attention}
\textbf{(GRCSA)}
It adopts a gated residual connection structure, where an input-driven gating network controls the strength of the residual connection. As shown in Figure \ref{fig4}(m), this method first sequentially applies channel attention and spatial attention to the input features in the manner of the classic CBAM to generate attention-enhanced features; then constructs a gating network to learn the gating value of the input features and constrains it between 0 and 1 via the sigmoid function; finally performs weighted fusion of the original input and the attention-enhanced features according to the gating value, where the weight of the original input is 1 minus the gating value, and the weight of the attention-enhanced features is the gating value itself. Through this gated residual design, the model can dynamically adjust the strength of the residual connection based on the adaptive characteristics of the input features, thereby achieving fine-grained residual-attention trade-offs across different samples or different spatial locations, and improving the model’s expressive capability and training stability. It is formulated as follows:
\begin{equation}
    W_{(m)} = \text{GA}(\textbf{X}_{(m)}^{in}),
\end{equation}
\begin{equation}
    \mathbf{X}_{(m)}^{out} = \textbf{X}_{(m)}^{in} \odot ( 1 -W_{(m)} ) + \text{SA}(\text{CA}(\textbf{X}_{(m)}^{in})) \odot W_{(m)}.
\end{equation}

\subsection{Multi-scale Information Mode}

\begin{figure}[H]
    \centering
    \includegraphics[width=0.8\linewidth]{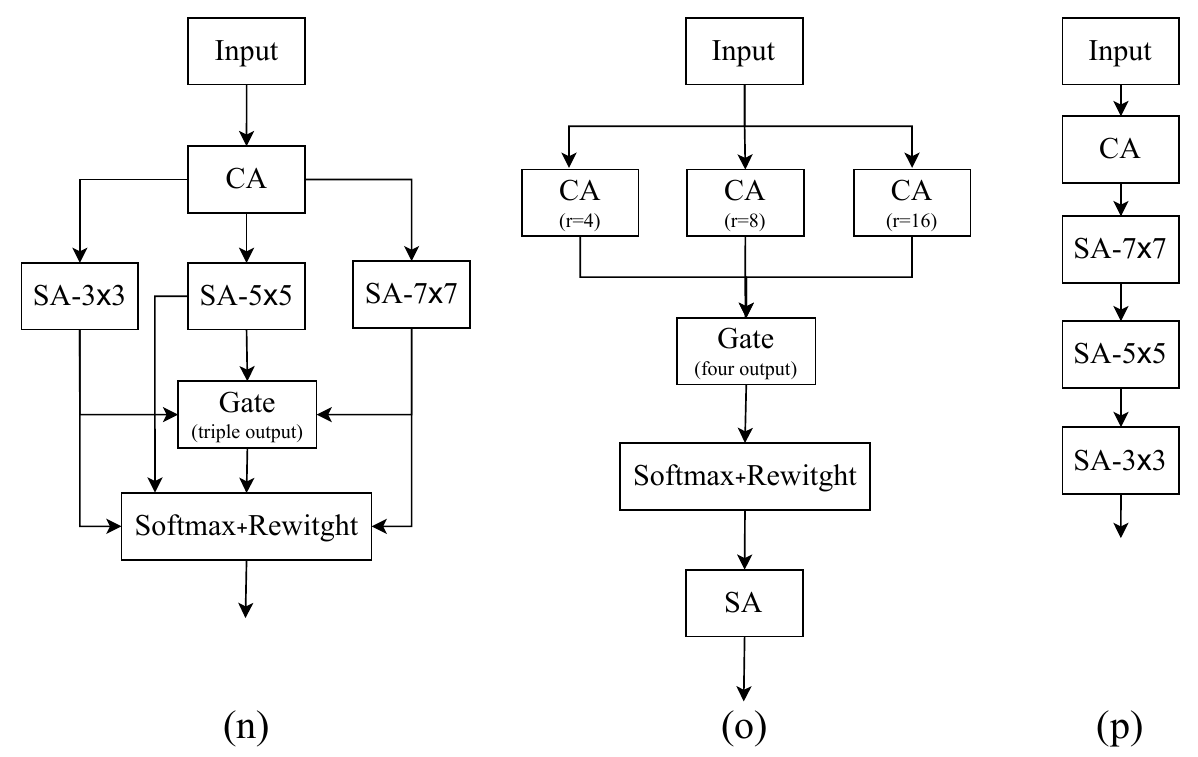}
    \caption{Structure Diagram of Multi-scale Information Pattern}
    \label{fig3}
\end{figure}

\textbf{Channel–Multi-Scale Spatial Attention (C-MSSA)}
It adopts a multi-scale parallel structure, as shown in Figure \ref{fig3}(n). After channel attention enhancement, multiple spatial attention modules with different convolution kernel sizes are applied to capture local, medium-scale, and global spatial dependencies respectively; then a gating network is constructed to learn adaptive fusion weights for the three multi-scale spatial attention modules; finally, the three weights are normalized using the softmax function, and the outputs of the three spatial attention branches are weighted and fused according to the normalized weights. Through this multi-scale design, the model can simultaneously focus on spatial features in different receptive field ranges, thereby achieving better recognition capability in scenarios with diverse target scales. It is formulated as follows:
\begin{equation}
\begin{split}
    \mathbf{X}_{(n)}^{1} &= \text{SA}_{3\times3}\bigl( \text{CA}(\mathbf{X}_{(n)}^{in}) \bigr), \\
    \mathbf{X}_{(n)}^{2} &= \text{SA}_{5\times5}\bigl( \text{CA}(\mathbf{X}_{(n)}^{in}) \bigr), \\
    \mathbf{X}_{(n)}^{3} &= \text{SA}_{7\times7}\bigl( \text{CA}(\mathbf{X}_{(n)}^{in}) \bigr),
\end{split}
\end{equation}
\begin{equation}
    \{\mathbf{W}_{(n)}^{1},\mathbf{W}_{(n)}^{2},\mathbf{W}_{(n)}^{3}\}= \text{Softmax(Gate}(
    \mathbf{X}_{(n)}^{1},     \mathbf{X}_{(n)}^{2},   \mathbf{X}_{(n)}^{3}
    )),
\end{equation}
\begin{equation}
      \mathbf{X}_{(n)}^{out}=
     W_{(n)}^{1} \odot  \mathbf{X}_{(n)}^{1} + 
     W_{(n)}^{2} \odot  \mathbf{X}_{(n)}^{2} + 
     W_{(n)}^{3} \odot  \mathbf{X}_{(n)}^{3}.
\end{equation}

\textbf{Multi-Squeeze Channel–Spatial Attention (MSC-SA)}
It applies multiple channel attention modules with different squeeze ratios for adaptive fusion. As shown in Figure \ref{fig3}(o), this method first applies three channel attention modules with different squeeze ratios in parallel to the input features to generate channel attention weights under different squeeze intensities respectively; then constructs a gating network to learn adaptive fusion weights for the three multi-squeeze channel attention modules; next normalizes the three weights using the softmax function, and performs weighted fusion of the outputs of the three channel attention branches according to the normalized weights; finally inputs the fused channel-enhanced features into the spatial attention module to generate spatial attention weights and complete the final enhancement. Through this multi-squeeze channel design, the model can achieve multi-scale trade-offs in channel squeeze intensity, thereby striking a better balance between information fidelity and compression efficiency. It is formulated as follows:
\begin{equation}
\begin{split}
    \mathbf{X}_{(o)}^{1} &= \mathrm{CA}_{r=4}\bigl( \mathbf{X}_{(o)}^{\mathrm{in}} \bigr), \\
    \mathbf{X}_{(o)}^{2} &= \mathrm{CA}_{r=8}\bigl( \mathbf{X}_{(o)}^{\mathrm{in}} \bigr), \\
    \mathbf{X}_{(o)}^{3} &= \mathrm{CA}_{r=16}\bigl( \mathbf{X}_{(o)}^{\mathrm{in}} \bigr),
\end{split}
\end{equation}
\begin{equation}
\{\mathbf{W}_{(o)}^{1},\mathbf{W}_{(o)}^{2},\mathbf{W}_{(o)}^{3}\}
= \mathrm{Softmax}\!\left(
\mathrm{Gate}\bigl(\mathbf{X}_{(o)}^{1},\mathbf{X}_{(o)}^{2},\mathbf{X}_{(o)}^{3}\bigr)
\right),
\end{equation}
\begin{equation}
\mathbf{X}_{(o)}^{\mathrm{out}} =
\mathrm{SA}\!\left(
\sum\nolimits_{k=1}^{3}\mathbf{X}_{(o)}^{k}\odot\mathbf{W}_{(o)}^{k}
\right).
\end{equation}

\textbf{Channel-Cascaded Multi-Scale Spatial Attention (C-CMSSA)}
As shown in Figure \ref{fig3}(p), this method first performs channel attention processing on the input features, and then sequentially applies spatial attention modules with large, medium, and small convolution kernels to the channel-enhanced features to capture coarse-scale, medium-scale, and fine-scale spatial dependencies respectively, achieving cascaded processing from coarse to fine. Through this cascaded multi-scale design, the model can gradually refine spatial structures within the progressively expanded effective receptive field, thereby achieving better performance in target recognition tasks with significant scale variations. It is formulated as follows:
\begin{equation}
    \mathbf{X}_{(p)}^{out} =
    \text{SA}_{3\times3}(
    \text{SA}_{5\times5}(
    \text{SA}_{7\times7}(\text{CA}(\textbf{X}_{(p)}^{in}))).
\end{equation}

\section{Experiments}
\label{sec:experiments}
\subsection{Datasets}
To comprehensively evaluate the generalization ability and task adaptability of attention mechanisms, this study constructs a heterogeneous benchmark testing system covering natural images and medical images, which includes 2 general visual datasets and 9 MedMNIST 2D medical datasets. These datasets span a sample size range from 9,000 to 107,000, a classification complexity from 2 to 14 categories, and differences in single-channel/three-channel modalities.

\textbf{CIFAR-10 Dataset}\cite{krizhevsky2009learning} consists of 60,000 color images with a resolution of 32×32 pixels, covering 10 categories—airplane, automobile, bird, cat, deer, dog, frog, horse, ship, and truck—and the sample size is evenly distributed across all categories.

\textbf{CIFAR-100 Dataset}\cite{krizhevsky2009learning} consists of 60,000 color images with a resolution of 32×32. This dataset adopts a hierarchical structure, organizing 100 fine-grained categories into 20 superclasses, where each superclass contains 5 subcategories. It provides rich semantic hierarchical information for fine-grained classification tasks. The specific superclasses are as follows: aquatic mammals, fish, flowers, food containers, fruits and vegetables, household appliances, furniture, insects, large carnivores, man-made outdoor scenes, natural outdoor scenes, large omnivores, medium-sized mammals, non-insect invertebrates, humans, reptiles, small mammals, trees, vehicles, and large man-made objects.

\textbf{MedMNIST2D} medical imaging library \cite{medmnistv2} provides high-quality medical data with unified specifications, standardized formats, and pre-divided training/validation/test sets, where all images have a resolution of 64×64 pixels. This characteristic significantly reduces the technical threshold for cross-study horizontal comparisons. To systematically explore the generalization behavior of attention mechanisms across different medical modalities, task complexities, and data scales, this study selects 9 representative tasks from more than ten of its sub-datasets, covering three major clinical modalities: pathology, dermatology, and radiology. Details of these datasets are shown in Table \ref{tab:medmnist2d}. The sample sizes span three orders of magnitude, and the number of categories ranges from 2 to 14, collectively forming a multi-level evaluation system from sparse small-sample to dense large-scale, and from binary classification to multi-class fine-grained classification.

\begin{table}[h]
\small
\setlength{\tabcolsep}{4pt}          
\caption{MedMNIST2D Dataset Summary (Grouped by Sample Size)}%
\label{tab:medmnist2d}
\begin{tabular*}{\linewidth}{@{\extracolsep{\fill}}lccc@{}}
\toprule
\textbf{Dataset} & \textbf{Data Modality} & \textbf{Tasks} & \textbf{Samples}\\
\midrule
BreastMNIST   & Breast Ultrasound     & Binary-Class (2)        & 780\\
RetinaMNIST   & Fundus Camera         & Ordinal Regression (5) & 1\,600\\
PneumoniaMNIST& Chest X-Ray           & Binary-Class (2)        & 5\,856\\
DermaMNIST    & Dermatoscope          & Multi-Class (7)         & 10\,015\\
BloodMNIST    & Blood Cell Microscope & Multi-Class (8)         & 17\,092\\
OrganCMNIST   & Abdominal CT          & Multi-Class (11)        & 23\,583\\
OrganSMNIST   & Abdominal CT          & Multi-Class (11)        & 25\,211\\
OrganAMNIST   & Abdominal CT          & Multi-Class (11) & 58\,830\\
PathMNIST     & Colon Pathology       & Multi-Class (9)         & 107\,180\\
\botrule
\end{tabular*}

\end{table}

\subsection{Implementation Details}

\textbf{Experiment Setting}
This study adopts the SGD optimizer with a momentum coefficient of 0.9 and a weight decay of $5 \times 10^{-4}$. The initial learning rate is uniformly set to 0.1, and the ReduceLROnPlateau scheduling strategy is employed: the learning rate is multiplied by 0.85 when the validation set accuracy does not improve for 5 consecutive epochs. Training configurations are dynamically adjusted according to different dataset scales: CIFAR-10 is trained for 150 epochs with a batch size of 256; for CIFAR-100, due to the increased number of categories to 100 and higher task complexity, the training epochs are extended to 300 while the batch size remains 256; the MedMNIST series datasets are uniformly trained for 100 epochs with a batch size of 128. To alleviate overfitting, label smoothing (label smoothing=0.1) is applied to CIFAR-100, and automatic calculation of class weights is optional to address data imbalance. Mixed Precision (AMP) training is enabled for acceleration, and the gradient clipping threshold is set to 0.5 to prevent gradient explosion. All experiments are repeated 3 times with a fixed random seed of 42, and the mean and standard deviation are reported to ensure result reliability. The experimental environment is based on the Linux system, Python 3.8 and PyTorch 2.0.0 framework, accelerated by NVIDIA RTX 3090 GPU and CUDA 11.8. VGG16 is selected as the backbone network in this study mainly based on three considerations: its regular structure, strong benchmarking capability, and computational efficiency. VGG16 has a clear hierarchical structure and distinct stages, facilitating the unified insertion of attention modules and ensuring the consistency of comparative experiments; as a widely validated classic architecture, it enables fair comparison with existing attention methods and improves the reproducibility and credibility of results; meanwhile, its moderate number of parameters supports large-scale attention combination experiments with limited resources, making it suitable for systematic exploration of various topological structures.

\begin{figure*}
    \centering
    \includegraphics[width=\linewidth]{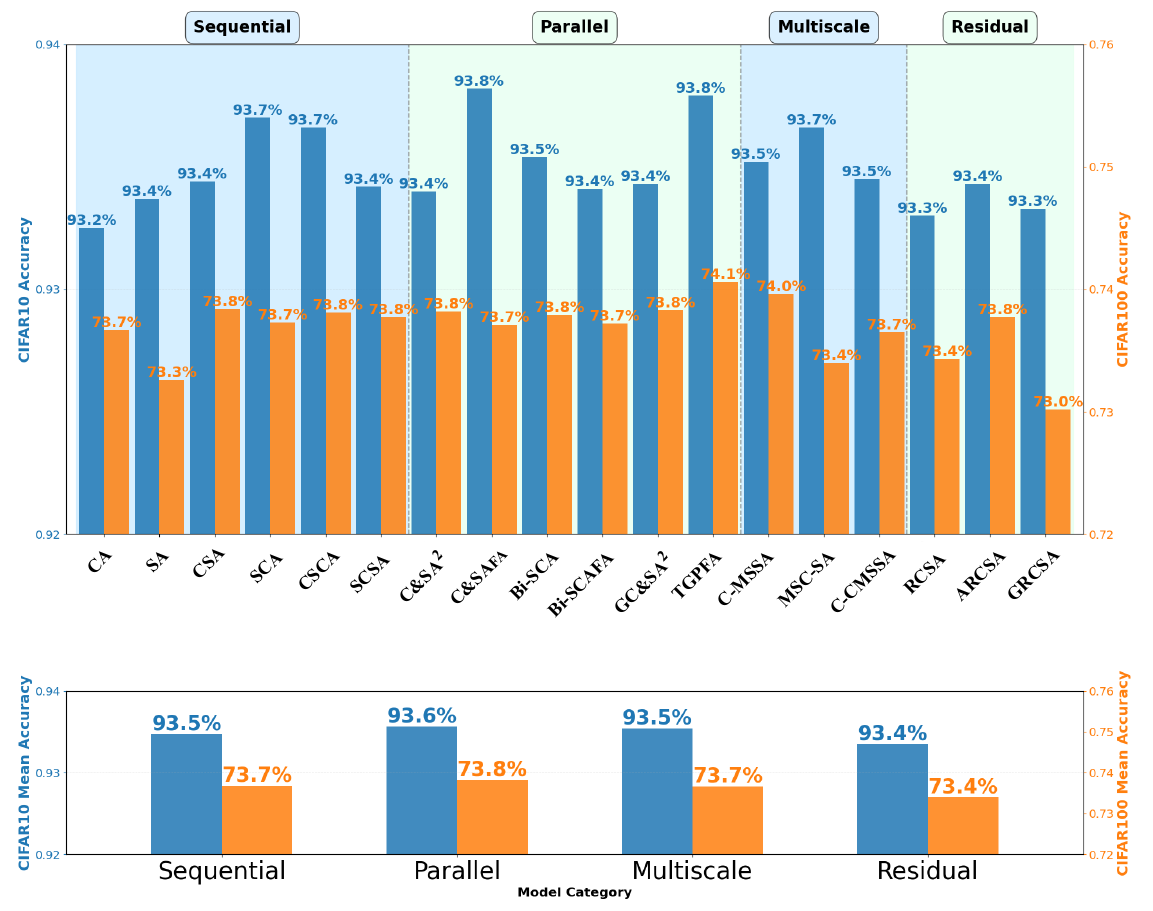}
    \caption{Validation accuracy comparison on CIFAR-10 (blue bars, left axis) and CIFAR-100 (orange bars, right axis) across 19 model modes. The highest accuracy for each dataset is labeled at the top of the corresponding bar.}
    \label{fig:cifar10_100}
\end{figure*}

\textbf{Evaluation Metrics}
This study adopts the following two core evaluation metrics to systematically assess the classification performance of different attention mechanism models on various datasets:

(1) \textbf{Cross-Entropy Loss}: It is used to measure the difference between the model's predicted probability distribution and the true label distribution, with a smaller value indicating more accurate model predictions. Its calculation formula is:
\begin{equation}
\mathcal{L}_{\text{CE}} = -\frac{1}{\text{N}} \sum_{i=1}^{\text{N}} \sum_{c=1}^{\text{C}} y_{ic} \log(p_{ic}),
\end{equation}
Among them, $\text{N}$ denotes the total number of samples in a batch, $\text{C}$ is the total number of categories, $y_{ic} \in \{0,1\}$ represents the true label of the $i$-th sample for category $c$, and $p_{ic}$ is the predicted probability output by the model that the $i$-th sample belongs to category $c$.

(2) \textbf{Classification Accuracy}: It is used to measure the proportion of samples correctly predicted by the model among the total samples, and is a core indicator for evaluating the overall performance of the model. Its calculation formula is:
\begin{equation}
\text{Accuracy} = \frac{\text{N}_{correct}}{\text{N}} = \frac{1}{\text{N}} \sum_{i=1}^{\text{N}} \text{I}\left(\arg\max_{c} p_{ic} = y_i\right),
\end{equation}
Among them, $\text{N}_{correct}$ denotes the total number of correctly predicted samples, $\text{N}$ is the total number of samples, $p_{ic}$ is the predicted probability output by the model that the $i$-th sample belongs to category $c$, $y_i$ is the true category label of the $i$-th sample, and $I(\cdot)$ is an indicator function which takes the value of 1 when the predicted category is consistent with the true category, and 0 otherwise.

\subsection{Comparison Experiments}
The experimental results on the two general visual benchmarks of CIFAR-10 and CIFAR-100 (Figure~\ref{fig:cifar10_100}) show that parallel attention mechanisms generally outperform serial structures. This performance confirms the advantage of parallel design in more fully capturing complementary multi-dimensional contextual information of channels and spaces. Among them, $\text{C} \& \text{SAFA}$ and $\text{TGPFA}$ achieve validation accuracies of 93.82\% and 93.79\% respectively on CIFAR-10, which are significantly higher than CBAM’s 93.44\%. In serial structures, the "spatial-first then channel" order achieves an accuracy of 93.70\%, which is also better than the traditional "channel-first then spatial" order, suggesting that prioritizing spatial region selection can provide a more effective feature basis for channel weighting. On CIFAR-100, a more challenging dataset for fine-grained classification tasks, the ternary parallel adaptive fusion strategy of $\text{TGPFA}$ achieves the highest accuracy. Its design of adaptively weighting and fusing original features, channel-enhanced features, and spatial-enhanced features can better capture subtle semantic differences between fine-grained categories.

Further observation of the MedMNIST medical imaging collection reveals that data scale exerts a significant moderating effect on attention mechanisms:

\begin{enumerate}
  \item \textbf{Small-Scale Binary Classification Tasks}
  On the BreastMNIST binary classification dataset of breast ultrasound images containing 780 samples, the cascaded multi-scale sequential attention structure of C-CMSSA achieves the best performance with an accuracy of 96.15\%, and the results are detailed in Table \ref{tab:combined_attention}. This indicates that in scenarios with extremely scarce data, multi-granularity spatial localization is more critical than enhancing channel semantics. The limited amount of data makes it difficult to support the stable learning of channel dependencies, while multi-scale spatial attention can effectively focus on lesion regions based on structural features, thereby improving the classification accuracy.

  \item \textbf{Multi-Class Fine-Grained Classification Tasks}
  On the DermaMNIST dataset of 7 classes of dermoscopic images with 10,015 samples, the parallel mechanism of $\text{C} \& \text{SAFA}$ significantly improves the accuracy from the baseline of 66.90\% to 81.06\%, as detailed in Table \ref{tab:parallel_attention}. Dermoscopic images are characterized by small inter-class differences and large intra-class variations. The learnable gating units in $\text{C}\&\text{SAFA}$ can adaptively balance the contribution weights of channel and spatial attention. This enables the model to simultaneously highlight discriminative channels (e.g., those related to lesion textures) and key spatial regions (e.g., lesion boundaries), effectively alleviating the confusion caused by intra-class variations.

  \item \textbf{Large-Scale Multi-Class Tasks}
   On the PathMNIST dataset, which contains 107,180 samples and 9 classes of colon pathological images, GC$\&$SA$^2$ achieves an accuracy of 99.57\% by dynamically gating and fusing parallel channel and spatial attention mechanisms. Driven by input data, its soft selection mechanism can adaptively adjust the attention intensity through channel gates and spatial gates respectively under the support of sufficient data, thus fully learning fine pathological feature patterns and effectively capturing subtle differences among different pathological types. For details, refer to Table \ref{tab:parallel_attention}.
\end{enumerate}

\begin{table}
\centering
\footnotesize
\caption{Performance of serial attention mechanisms across all datasets}
\label{tab:serial_attention}
\begin{tabular}{lccccccc}
\toprule
\textbf{Dataset} & Baseline & CA & SA & CSA & SCA & CSCA & SCSA \\
\midrule
CIFAR-10 & 93.21 & 93.25 & 93.37 & 93.44 & 93.70 & 93.66 & 93.42 \\
CIFAR-100 & 73.61 & 73.67 & 73.26 & 73.84 & 73.73 & 73.81 & 73.77 \\
RetinaMNIST & 50.83 & 53.33 & 55.00 & 53.33 & 53.33 & 51.67 & 51.67 \\
PneumoniaMNIST & 98.66 & 99.05 & 98.86 & 99.24 & \textcolor{red}{99.62} & 99.05 & 99.43 \\
PathMNIST & 99.20 & 98.83 & 98.76 & 99.37 & 99.31 & 98.94 & 99.39 \\
OrgansMNIST & 92.25 & 93.31 & 92.94 & 93.35 & 92.94 & 92.99 & 93.19 \\
OrganCMNIST & 98.87 & 98.95 & 98.58 & 98.70 & 98.58 & 98.83 & 98.79 \\
DermaMNIST & 66.90 & 61.52 & 66.90 & 66.90 & 68.20 & 66.60 & 66.90 \\
BreastMNIST & 88.46 & 93.59 & 73.08 & 94.87 & 93.59 & 92.31 & 93.59 \\
BloodMNIST & 99.01 & 99.02 & 99.01 & 98.89 & 98.95 & 99.01 & \textcolor{red}{99.30} \\
OrganAMNIST & 99.78 & 99.46 & 99.35 & 99.54 & 99.51 & 99.12 & 99.48 \\
\bottomrule
\end{tabular}
\footnotesize{$\dagger$ Pairwise bootstrap t-test against baseline shows $p < 0.01$ for all top-3 models.}
\end{table}

\begin{table}
\centering
\footnotesize
\caption{Performance of parallel attention mechanisms across all datasets}
\label{tab:parallel_attention}
\begin{tabular}{lccccccc}
\toprule
\textbf{Dataset} & Baseline & C\&SA$^2$ & C\&SAFA & Bi-CSA & Bi-CSAFA & GC\&SA$^2$ & TGPFA \\
\midrule
CIFAR-10 & 93.21 & 93.40 & \textcolor{red}{93.82} & 93.54 & 93.41 & 93.43 & 93.79 \\
CIFAR-100 & 73.61 & 73.82 & 73.71 & 73.79 & 73.72 & 73.83 & \textcolor{red}{74.06} \\
RetinaMNIST & 50.83 & 53.33 & 55.00 & 54.17 & 50.00 & 55.00 & \textcolor{red}{55.83} \\
PneumoniaMNIST & 98.66 & 99.05 & 98.66 & 98.86 & 99.43 & 99.05 & 98.86 \\
PathMNIST & 99.20 & 99.48 & 99.16 & 99.23 & 99.29 & \textcolor{red}{99.57} & 99.18 \\
OrgansMNIST & 92.25 & 93.15 & 92.62 & 93.27 & 92.94 & 92.58 & 92.82 \\
OrganCMNIST & 98.87 & 98.70 & 98.37 & \textcolor{red}{99.00} & 98.87 & 98.54 & 98.95 \\
DermaMNIST & 66.90 & 66.10 & \textcolor{red}{81.06} & 66.90 & 70.29 & 63.31 & 65.90 \\
BreastMNIST & 88.46 & 92.31 & 92.31 & 91.03 & 96.14 & 91.03 & 88.46 \\
BloodMNIST & 99.01 & 99.01 & 99.07 & 99.24 & 98.83 & 99.01 & 98.77 \\
OrganAMNIST & 99.78 & 99.57 & 88.74 & 15.91 & 99.26 & \textcolor{red}{99.75} & 99.38 \\
\bottomrule
\end{tabular}
\footnotesize{$\dagger$ Pairwise bootstrap t-test against baseline shows $p < 0.01$ for all top-3 models.}
\end{table}

\begin{table}
\centering
\footnotesize
\caption{Performance of residual connection and multi-scale attention mechanisms across all datasets}
\label{tab:combined_attention}
\begin{tabular}{lccccccc}
\toprule
\multirow{2}{*}{\textbf{Dataset}} &
\multirow{2}{*}{Baseline} &
\multicolumn{3}{c}{\textbf{Residual Connection}} & 
\multicolumn{3}{c}{\textbf{Multi-scale}} \\
\cmidrule(lr){3-5} \cmidrule(lr){6-8}
& & RCSA & ARCSA & GRCSA & 
C-MSSA & MSC-SA & C-CMSSA \\
\midrule
CIFAR-10 & 93.21 & 93.30 & 93.43 & 93.33 & 93.52 & 93.66 & 93.45 \\
CIFAR-100 & 73.61 & 73.43 & 73.77 & 73.02 & 73.96 & 73.40 & 73.65 \\
RetinaMNIST & 50.83 & 52.50 & 51.67 & 54.17 & 53.33 & 54.17 & \textcolor{red}{55.83} \\
PneumoniaMNIST & 98.66 & 98.47 & 99.05 & 99.05 & 99.43 & 98.66 & 99.24 \\
PathMNIST & 99.20 & 99.17 & 99.31 & 98.89 & 99.37 & 99.19 & 98.18 \\
OrgansMNIST & 92.25 & 92.86 & 92.90 & 93.27 & 93.39 & \textcolor{red}{93.47} & 92.33 \\
OrganCMNIST & 98.87 & 98.66 & 98.95 & 98.87 & 98.58 & 98.54 & 98.16 \\
DermaMNIST & 66.90 & 66.90 & 71.59 & 67.70 & 67.50 & 68.30 & 69.39 \\
BreastMNIST & 88.46 & 91.03 & 92.31 & 92.31 & 92.31 & 93.59 & \textcolor{red}{96.15} \\
BloodMNIST & 99.01 & 98.77 & 99.12 & 99.12 & 99.01 & 98.66 & 99.01 \\
OrganAMNIST & 99.78 & 99.54 & 99.45 & \textcolor{red}{99.75} & 99.68 & 99.68 & 99.17 \\
\bottomrule
\end{tabular}
\footnotesize{$\dagger$ Pairwise bootstrap t-test against baseline shows $p < 0.01$ for all top-3 models.}
\end{table}

To evaluate the computational overhead, we sequentially integrated 18 attention topologies into the VGG backbone network and conducted tests in a unified environment. Table~\ref{tab:flops} shows that each module introduces only a minimal number of parameters and almost no additional FLOPs, fully verifying the lightweight characteristics of the proposed methods.

\begin{table}
\centering
\small
\caption{Computational overhead of 18 attention topologies. Input $64\!\times\!64$, batch size 1.}
\label{tab:flops}
\begin{tabular}{cccc}
\toprule
Category & Attention mechanism & FLOPs (G) & Params (M) \\
\midrule
\multirow{6}{*}{Serial}
& VGG + CA   & 0.314 & 15.069 \\
& VGG + SA   & 0.314 & 14.991 \\
& VGG + CSA  & 0.314 & 15.069 \\
& VGG + SCA  & 0.314 & 15.069 \\
& VGG + CSCA & 0.314 & 15.069 \\
& VGG + SCSA & 0.314 & 15.069 \\
\midrule
\multirow{6}{*}{Parallel}
& VGG + C\&SA$^2$   & 0.314 & 14.991 \\
& VGG + C\&SAFA    & 0.314 & 14.991 \\
& VGG + Bi-CSA     & 0.314 & 14.991 \\
& VGG + Bi-CSAFA   & 0.314 & 14.991 \\
& VGG + TGPFA      & 0.314 & 14.991 \\
& VGG + GC\&SA$^2$ & 0.314 & 15.069 \\
\midrule
\multirow{3}{*}{Residual}
& VGG + RCSA  & 0.314 & 14.991 \\
& VGG + ARCSA & 0.314 & 14.991 \\
& VGG + GRCSA & 0.314 & 14.991 \\
\midrule
\multirow{3}{*}{Multi-scale}
& VGG + C-MSSA   & 0.314 & 15.069 \\
& VGG + MSC-SA   & 0.314 & 15.069 \\
& VGG + C-CMSSA  & 0.314 & 15.069 \\
\bottomrule
\end{tabular}
\end{table}

\subsection{t-SNE Visualization}
\begin{figure*}
    \centering
    \includegraphics[width=\linewidth]{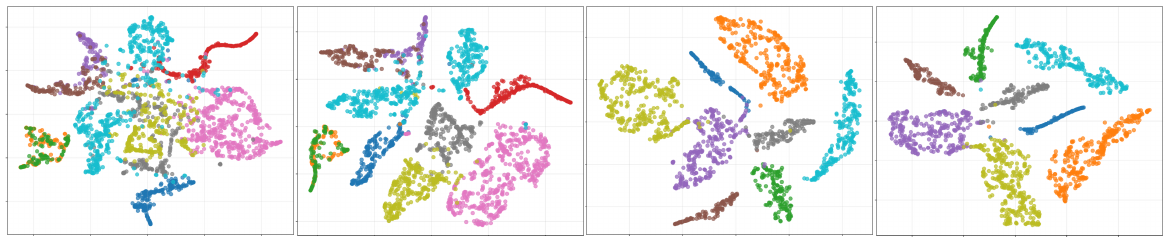}
    \caption{T-SNE visualization. The two figures on the left show VGG and VGG + MSC-SA on the OrganSMNIST dataset; the two figures on the right show VGG and VGG + SCSA on the BloodMNIST dataset}
    \label{fig:T-SNE}
\end{figure*}

t-SNE (t-distributed Stochastic Neighbor Embedding) is an algorithm that reduces high-dimensional features to two dimensions while preserving local neighborhood relationships, which can intuitively demonstrate the intra-class aggregation and inter-class separation of features learned by the network. To qualitatively verify the effectiveness of attention modules, this paper visualizes the penultimate layer features of two representative medical datasets. Figure~\ref{fig:T-SNE} compares the VGG backbone with the optimal attention configurations: MSC-SA is adopted on OrganSMNIST (11 classes), and SCSA is adopted on BloodMNIST (8 classes). The results show that the cluster structures generated by the attention models are more compact and the category boundaries are clearer, which is consistent with the improvement of quantitative metrics.

\section{Discussion}
\textbf{Sequential vs. Parallel Structures: The Impact of Data Volume on Selection}
Through experiments, we found that as the volume of training data increases, the optimal attention combination pattern shifts from "sequential structure → parallel structure → parallel structure with gating" rather than a fixed superior structure. When the number of samples is very small (i.e.,$N<1k$), sequential combinations like C-CMSSA with "channel-first followed by multi-scale spatial attention" are more suitable. It first reduces useless channel information and lowers the computational load of the spatial attention component, thereby avoiding model overfitting. When the sample size is medium (i.e., $1k \le N \le 50k$), the data is sufficient to support the model in learning fusion rules, making parallel structures such as C$\&$SAFA and Bi-CSAFA more advantageous. The two branches process channel and spatial information separately, complementing each other’s deficiencies and enabling the model to learn more comprehensive features. When the sample size is large (i.e., $N>50k$), parallel structures with dynamic gating like GC$\&$SA$^2$ perform best. Abundant data allows the model to precisely adjust the attention intensity of channels and spaces, and each sample can obtain exclusive feature optimization—this advantage of flexible adaptation surpasses the limitations of sequential structures. It can thus be concluded that the claim "parallel is always better than sequential" is invalid; the key lies in selection based on data volume.

\textbf{Order of Sequential Structures: Spatial-First Then Channel Is More Stable}
On the CIFAR-10 and CIFAR-100 datasets, the Spatial-then-Channel Attention (SCA) achieves a stably higher accuracy of 0.2\%-0.3\% compared with the traditional Channel-then-Spatial Attention (CSA). Spatial attention first focuses on the key positions of the image and completely preserves these important structures; then channel attention selects meaningful channel information from these features with "already identified positions", eliminating the need to simultaneously consider "locating positions" and "selecting channels", which makes it easier for the model to learn effectively. Conversely, if channel attention is applied first, it may suppress some channels that seem unimportant but contain key details, and it will be difficult for subsequent spatial attention to retrieve these details, leading to ambiguous boundaries for classification judgments. This "first locate positions, then distinguish meanings" approach is consistent with the logic of human observation of things, and its advantages are obvious in classification tasks that require distinguishing subtle differences.

\textbf{Role of Residual Connections: Simple Addition Is More Reliable, Complex Adjustment Depends on Data Volume}
The core advantage of residual connections is transforming the "multiplicative weighting" of attention modules into "additive supplementation", which avoids the problems of the model failing to learn effective information and gradient transmission interruption when attention weights are too small.
Experiments show that simply adding a channel for "direct transmission of original features" to the attention module (i.e., RCSA) can stably improve the accuracy by 0.33\% on CIFAR-100. If the model is to learn the fusion ratio of original features and attention features by itself (e.g., ARCSA, GRCSA), the effect is related to data volume: with a large number of samples, GRCSA can flexibly adjust the fusion ratio according to the characteristics of each sample, enabling the model to learn more effective information and achieve better performance; with a small number of samples, such fusion parameters that need to be learned will instead become an "extra burden", leading to greater fluctuations in the model’s prediction results. It can thus be seen that "simple addition" is sufficient for small-sample scenarios, and adaptive gating is only necessary for large-sample scenarios.

\textbf{Multi-Scale Strategies: Selecting the Right Type to Match Task Characteristics}
Different multi-scale attention mechanisms are suitable for different types of tasks:
C-MSSA achieves the best performance on RetinaMNIST because lesions in fundus images vary significantly in size and location, requiring simultaneous attention to global and local information. This "spatial multi-scale" design can accurately capture lesions of different sizes.
MSC-SA performs better on BloodMNIST because the differences between different blood cells mainly lie in texture and color, which are more sensitive to the compression ratio of channel information. This "channel multi-scale" design can better distinguish these subtle differences.
This indicates that when selecting a multi-scale strategy, it is necessary to combine task characteristics: if the task requires detecting targets of different sizes, spatial multi-scale should be used; if the task requires distinguishing details such as texture and color of targets, channel multi-scale should be used.

\section{Conclusion}
This paper addresses the core problem of combination optimization of channel and spatial attention, systematically comparing 18 topological structures across four dimensions—sequential, parallel, residual, and multi-scale within a unified framework. Through cross-scenario experiments on two general vision datasets and nine MedMNIST medical imaging datasets, we draw the following conclusions: First, attention topology exhibits a "three-stage" coupling relationship with data scale, and there exists no absolutely optimal fixed combination. Second, within sequential structures, the "Spatial→Channel" order demonstrates superior performance in fine-grained classification tasks. Third, residual connections can alleviate gradient vanishing, while advanced designs incorporating learnable weights require sufficient data volume for support. Based on these findings, this paper proposes directly implementable design guidelines: prioritize "Channel→Multi-scale Spatial" cascaded structures for small-sample datasets; employ parallel structures with learnable fusion weights for datasets with sample sizes between 1k and 50k; incorporate input-driven dynamic gating mechanisms in parallel branches when sample size exceeds 50k; and for detail-sensitive tasks, adopt the "Spatial→Channel" order regardless of data scale, combined with residual connections. Future work will explore topological generalizability across different network backbones and more complex vision-language tasks, and further derive error bounds for different structures from an information-theoretic perspective to establish a complete attention design theory.

\section*{Funding}
None.

\bibliography{sn-bibliography}

@article{Transformer,
  title={Attention is all you need},
  author={Vaswani, Ashish and Shazeer, Noam and Parmar, Niki and Uszkoreit, Jakob and Jones, Llion and Gomez, Aidan N and Kaiser, {\L}ukasz and Polosukhin, Illia},
  journal={Advances in neural information processing systems},
  volume={30},
  year={2017}
}

@INPROCEEDINGS{SENet,
  author={Hu, Jie and Shen, Li and Sun, Gang},
  booktitle={2018 IEEE/CVF Conference on Computer Vision and Pattern Recognition}, 
  title={Squeeze-and-Excitation Networks}, 
  year={2018},
  volume={},
  number={},
  pages={7132-7141},
  doi={10.1109/CVPR.2018.00745}}

@inproceedings{CBAM,
  title={Cbam: Convolutional block attention module},
  author={Woo, Sanghyun and Park, Jongchan and Lee, Joon-Young and Kweon, In So},
  booktitle={Proceedings of the European conference on computer vision (ECCV)},
  doi={https://doi.org/10.1007/978-3-030-01234-2_1},
  pages={3--19},
  year={2018}
}

@inproceedings{BAM,
  title={BAM: Bottleneck Attention Module},
  author={Jongchan Park and Sanghyun Woo and Joon-Young Lee and In-So Kweon},
  booktitle={British Machine Vision Conference},
  year={2018},
}

@inproceedings{DANet,
  title={Dual attention network for scene segmentation},
  author={Fu, Jun and Liu, Jing and Tian, Haijie and Li, Yong and Bao, Yongjun and Fang, Zhiwei and Lu, Hanqing},
  booktitle={Proceedings of the IEEE/CVF conference on computer vision and pattern recognition},
  pages={3146--3154},
  year={2019}
}

@inproceedings{ECANet,
  title={ECA-Net: Efficient channel attention for deep convolutional neural networks},
  author={Wang, Qilong and Wu, Banggu and Zhu, Pengfei and Li, Peihua and Zuo, Wangmeng and Hu, Qinghua},
  booktitle={Proceedings of the IEEE/CVF conference on computer vision and pattern recognition},
  pages={11534--11542},
  year={2020}
}

@inproceedings{SKNet,
  title={Selective kernel networks},
  author={Li, Xiang and Wang, Wenhai and Hu, Xiaolin and Yang, Jian},
  booktitle={Proceedings of the IEEE/CVF conference on computer vision and pattern recognition},
  pages={510--519},
  year={2019}
}

@inproceedings{GCNet,
  title={Gcnet: Non-local networks meet squeeze-excitation networks and beyond},
  author={Cao, Yue and Xu, Jiarui and Lin, Stephen and Wei, Fangyun and Hu, Han},
  booktitle={Proceedings of the IEEE/CVF international conference on computer vision workshops},
  pages={0--0},
  year={2019}
}

@inproceedings{CoordinateAttention,
  title={Coordinate attention for efficient mobile network design},
  author={Hou, Qibin and Zhou, Daquan and Feng, Jiashi},
  booktitle={Proceedings of the IEEE/CVF conference on computer vision and pattern recognition},
  pages={13713--13722},
  year={2021}
}

@article{PSA,
  title={Polarized self-attention: Towards high-quality pixel-wise regression},
  author={Liu, Huajun and Liu, Fuqiang and Fan, Xinyi and Huang, Dong},
  journal={arXiv preprint arXiv:2107.00782},
  year={2021}
}

@article{li2019sge,
  title={Spatial group-wise enhance: Improving semantic feature learning in convolutional networks},
  author={Li, Xiang and Hu, Xiaolin and Yang, Jian},
  journal={arXiv preprint arXiv:1905.09646},
  year={2019}
}

@article{GCE,
  title={Multiscale sparse cross-attention network for remote sensing scene classification},
  author={Ma, Jingjing and Jiang, Wei and Tang, Xu and Zhang, Xiangrong and Liu, Fang and Jiao, Licheng},
  journal={IEEE Transactions on Geoscience and Remote Sensing},
  year={2025},
  publisher={IEEE}
}

@article{PCBAM,
  title={DAU-Net: Dual attention-aided U-Net for segmenting tumor in breast ultrasound images},
  author={Pramanik, Payel and Roy, Ayush and Cuevas, Erik and Perez-Cisneros, Marco and Sarkar, Ram},
  journal={Plos one},
  volume={19},
  number={5},
  pages={e0303670},
  year={2024},
  publisher={Public Library of Science San Francisco, CA USA}
}

@inproceedings{CS-Net,
  title={CS-Net: Channel and spatial attention network for curvilinear structure segmentation},
  author={Mou, Lei and Zhao, Yitian and Chen, Li and Cheng, Jun and Gu, Zaiwang and Hao, Huaying and Qi, Hong and Zheng, Yalin and Frangi, Alejandro and Liu, Jiang},
  booktitle={International Conference on Medical Image Computing and Computer-Assisted Intervention},
  pages={721--730},
  year={2019},
  organization={Springer}
}

@article{CHEN2025111958,
title = {An efficient multi-task forest fire and smoke detection model},
journal = {Engineering Applications of Artificial Intelligence},
volume = {160},
pages = {111958},
year = {2025},
author = {Cong Chen and Yunfei Liu and Chenyu Zhang and Junhui Li and Xingliang Chen}
}

@ARTICLE{GraphConvolutionAttention,
  author={Peng, Yali and Li, Hong and Wang, Meiyun and Qin, Le and Du, Yingkui and Yi, Yugen},
  journal={IEEE Journal of Biomedical and Health Informatics}, 
  title={PMSFINet: Progressive Multi-Scale Feature Interaction Network for Medical Image Segmentation}, 
  year={2025},
  volume={},
  number={},
  pages={1-14}
  }

@INPROCEEDINGS{Wazir,
  author={Wazir, Saad and Kim, Daeyoung},
  booktitle={2025 IEEE/CVF Conference on Computer Vision and Pattern Recognition (CVPR)},
  title={Rethinking Decoder Design: Improving Biomarker Segmentation Using Depth-to-Space Restoration and Residual Linear Attention}, 
  year={2025},
  volume={},
  number={},
  pages={30861-30871}}

@InProceedings{Rahman_2024_CVPR,
    author    = {Rahman, Md Mostafijur and Munir, Mustafa and Marculescu, Radu},
    title     = {EMCAD: Efficient Multi-scale Convolutional Attention Decoding for Medical Image Segmentation},
    booktitle = {Proceedings of the IEEE/CVF Conference on Computer Vision and Pattern Recognition (CVPR)},
    month     = {June},
    year      = {2024},
    pages     = {11769-11779}
}

@techreport{krizhevsky2009learning,
  title={Learning multiple layers of features from tiny images},
  author={Krizhevsky, Alex and Hinton, Geoffrey},
  institution={University of Toronto},
  year={2009}
}

@article{medmnistv2,
    title={MedMNIST v2-A large-scale lightweight benchmark for 2D and 3D biomedical image classification},
    author={Yang, Jiancheng and Shi, Rui and Wei, Donglai and Liu, Zequan and Zhao, Lin and Ke, Bilian and Pfister, Hanspeter and Ni, Bingbing},
    journal={Scientific Data},
    volume={10},
    number={1},
    pages={41},
    year={2023},
    publisher={Nature Publishing Group UK London}
}

@INPROCEEDINGS{11239364,
  author={Zhongming, Liu and Huang, Xin and Li, Xiao and Zou, Xiang},
  booktitle={2025 19th International Conference on Complex Medical Engineering (CME)}, 
  title={SGMSNet: Spatial-Gated Memory Network for Chest X-ray Image Segmentation}, 
  year={2025},
  volume={},
  number={},
  pages={111-115},
  doi={10.1109/CME67420.2025.11239364}}

@article{FENG2026132232,
title = {Smart CSWin-UNet: Integrating prototype attention gate and mixture-of-experts skip connections for medical image segmentation},
journal = {Neurocomputing},
volume = {665},
pages = {132232},
year = {2026},
issn = {0925-2312},
doi = {https://doi.org/10.1016/j.neucom.2025.132232},
author = {Chuanbo Feng and Xinchu Lu and Jing Qin and Daoqiang Zhang and Xiaoke Hao},
}

@article{ELSAABRAHAM2025132379,
title = {Saliency-guided contrastive pretraining and attention-enhanced decoding for semi-supervised diabetic retinopathy lesion segmentation},
journal = {Neurocomputing},
pages = {132379},
year = {2025},
issn = {0925-2312},
doi = {https://doi.org/10.1016/j.neucom.2025.132379},
author = {Shilpa {Elsa Abraham} and Binsu C. Kovoor and S. Nidhi},
}

@article{ZHANG2025132338,
title = {PolySAGN: Hierarchical multi-scale representation learning with scale-specific attention for accurate polyp segmentation},
journal = {Neurocomputing},
pages = {132338},
year = {2025},
issn = {0925-2312},
doi = {https://doi.org/10.1016/j.neucom.2025.132338},
author = {Wenqi Zhang and Yue Zhang and Muhammad Fayaz and L.Minh Dang and Tan N. Nguyen and Hyeonjoon Moon}
}

@article{LIU2025131369,
title = {FANCL: Feature-guided attention network with curriculum learning for brain metastases segmentation},
journal = {Neurocomputing},
volume = {655},
pages = {131369},
year = {2025},
issn = {0925-2312},
doi = {https://doi.org/10.1016/j.neucom.2025.131369},
author = {Zijiang Liu and Xiaoyu Liu and Linhao Qu and Yonghong Shi}
}

@article{ZHAO2025131101,
title = {Dual-stage learning framework for underwater acoustic target recognition with cross-attention mechanism and audio-guided contrastive learning},
journal = {Neurocomputing},
volume = {652},
pages = {131101},
year = {2025},
issn = {0925-2312},
doi = {https://doi.org/10.1016/j.neucom.2025.131101},
author = {Rongyao Zhao and Feng Liu and Lyufang Zhao and Daihui Li and Jing Xu and Yuanxin Liu and Tongsheng Shen}
}

@article{LV2026109537,
title = {Simulation of the coupled navier-stokes-cahn-hilliard-heat transfer system via a U-shaped spatial-channel attention neural operator},
journal = {Communications in Nonlinear Science and Numerical Simulation},
volume = {154},
pages = {109537},
year = {2026},
issn = {1007-5704},
doi = {https://doi.org/10.1016/j.cnsns.2025.109537},
author = {Zhixian Lv and Yuhong Li and Qing Xia and Junseok Kim and Yibao Li},
}

@article{JI2026113141,
title = {A Point-Voxel Transformer for point cloud object detection with spatial and channel attention},
journal = {Engineering Applications of Artificial Intelligence},
volume = {163},
pages = {113141},
year = {2026},
issn = {0952-1976},
doi = {https://doi.org/10.1016/j.engappai.2025.113141},
author = {Guangyu Ji and Jun Lu and Chengtao Cai and Kaibin Qin},
}

@article{MA2025,
title = {Irregularly seismic data interpolation based on deep learning with integrated channel-spatial attention mechanism},
journal = {Petroleum Science},
year = {2025},
issn = {1995-8226},
doi = {https://doi.org/10.1016/j.petsci.2025.10.004},
author = {Chao Ma and Jian-Ping Huang and Zi-Xuan Qiao and San-Fu Li and Wen-Sheng Duan and Gang-Lin Lei},
}

@article{WU2026105770,
title = {Lightweight multi-scale dynamic feature focusing network integrating spatial channel attention mechanism for autonomous driving object detection},
journal = {Digital Signal Processing},
volume = {170},
pages = {105770},
year = {2026},
issn = {1051-2004},
doi = {https://doi.org/10.1016/j.dsp.2025.105770},
author = {Kunpeng Wu and Yunfeng Xu and Junpo Zhang},

}

@article{PENG2026108807,
title = {SCEA-Net: A hybrid framework from spatial-channel-aware external attention for accurate 3D medical image segmentation},
journal = {Biomedical Signal Processing and Control},
volume = {113},
pages = {108807},
year = {2026},
issn = {1746-8094},
doi = {https://doi.org/10.1016/j.bspc.2025.108807},
author = {Gennian Peng and Xuesong Lu and Yong Chen and Hong Chen and Zhiwei Zhai and Tong Chen and Qinlan Xie},
}

@article{VENTURINI2025105754,
title = {Leveraging spatial-channel attention in U-Net for enhanced segmentation of martian dust storms},
journal = {Image and Vision Computing},
volume = {163},
pages = {105754},
year = {2025},
issn = {0262-8856},
doi = {https://doi.org/10.1016/j.imavis.2025.105754},
author = {Daniele Venturini and Marco Raoul Marini and Luigi Cinque and Gian Luca Foresti},
}

@article{LU2026110288,
title = {CM-CSAMFNet: A cross-modality channel and spatial attention module fusion network for multimodal medical image fusion},
journal = {Signal Processing},
volume = {239},
pages = {110288},
year = {2026},
issn = {0165-1684},
doi = {https://doi.org/10.1016/j.sigpro.2025.110288},
author = {Yixiang Lu and Changqing Xu and Jingyun Gong and Qingwei Gao and Dong Sun and De Zhu}
}

@article{LI2026108243,
title = {Robust medical image zero watermarking algorithm based on Spatial and Channel Synergistic Attention and FocalNet},
journal = {Biomedical Signal Processing and Control},
volume = {112},
pages = {108243},
year = {2026},
issn = {1746-8094},
doi = {https://doi.org/10.1016/j.bspc.2025.108243},
author = {Jingyou Li and Fengshan Zhang and Guangda Zhang and Zixin Yang and Haiyang Wei},
}

@article{SI2025129866,
title = {SCSA: Exploring the synergistic effects between spatial and channel attention},
journal = {Neurocomputing},
volume = {634},
pages = {129866},
year = {2025},
issn = {0925-2312},
doi = {https://doi.org/10.1016/j.neucom.2025.129866},
author = {Yunzhong Si and Huiying Xu and Xinzhong Zhu and Wenhao Zhang and Yao Dong and Yuxing Chen and Hongbo Li}
}

@article{FANG2025107474,
title = {A diagnosis and monitoring system with a multi-scale channel and spatial attention mechanism on residual structures for tubing leakage detection},
journal = {Process Safety and Environmental Protection},
volume = {201},
pages = {107474},
year = {2025},
issn = {0957-5820},
doi = {https://doi.org/10.1016/j.psep.2025.107474},
author = {Yilin Fang and Jianchun Fan and Yunpeng Yang and Fanfan Ma and Guoqing Ren},

}

@article{ZHAO2025111520,
title = {MSCIAIG-Net: Multi-scale spatial-channel interactive attention and interpretable guidance for fine-grained ship classification and visual inference network in remote sensing images},
journal = {Engineering Applications of Artificial Intelligence},
volume = {157},
pages = {111520},
year = {2025},
issn = {0952-1976},
doi = {https://doi.org/10.1016/j.engappai.2025.111520},
author = {Chengqiang Zhao and Shijie Chen and Jiashu Zhang}
}

@article{CAI2025114285,
title = {CT-SSSA: Malicious traffic augmentation based on classifier transGAN and spatial-channel synergistic self-attention},
journal = {Knowledge-Based Systems},
volume = {329},
pages = {114285},
year = {2025},
issn = {0950-7051},
doi = {https://doi.org/10.1016/j.knosys.2025.114285},
author = {Saihua Cai and Xingyu Zhao and Jinfu Chen and Yige Zhao and Junyi Chen and Lizhou Chen},

}

@article{WANG2025111907,
title = {Loosen Attention: Integrating localized channel and coarse spatial attention for enhanced analysis of complex aurora images},
journal = {Engineering Applications of Artificial Intelligence},
volume = {160},
pages = {111907},
year = {2025},
issn = {0952-1976},
doi = {https://doi.org/10.1016/j.engappai.2025.111907},
author = {Qian Wang and Shihao Jing and Rui Yang and Zhenpei Liu and Yao Tang and Han Pan}
}

\end{document}